%% file: bare_jrnl_new_sample4.tex
\documentclass[lettersize,journal]{IEEEtran}
\usepackage{amsmath,amsfonts}
\usepackage{algorithmic}
\usepackage{algorithm}
\usepackage{array}
\usepackage{textcomp}
\usepackage{url}
\usepackage{verbatim}
\usepackage{graphicx}
\usepackage{subcaption}
\usepackage{cite}
\usepackage{xspace}
\usepackage{xcolor}
\usepackage{amsmath,amssymb}
\usepackage{booktabs}
\usepackage{multirow}
\usepackage{tcolorbox}
\usepackage[normalem]{ulem}
\usepackage{dblfloatfix}
\usepackage{tikz}
\usepackage{pgfplots}
\pgfplotsset{compat=1.18}
\usepgfplotslibrary{statistics}
\usepgfplotslibrary{colormaps}

\usepackage{hyperref}

\usepackage{orcidlink}

\begin{document}


\title{Do Input-Level Defenses Transfer to Observation-Level Attacks on VideoLLMs?}

\author{
Bangshuo Zhu$^{\orcidlink{0009-0000-5451-6166}}$,
Wei Song$^{\orcidlink{0000-0002-1240-8412}}$,
Yuxin Cao$^{\orcidlink{0009-0002-5766-0846}}$,
Yuezhong Wu$^{\orcidlink{0000-0002-0866-5379}}$,
Zhiquan Liu$^{\orcidlink{0000-0002-3934-2177}}$,
Yuekang Li$^{\orcidlink{0000-0003-0237-4658}}$,
Jingling Xue$^{\orcidlink{0000-0003-0380-3506}}$

\IEEEcompsocitemizethanks{
\IEEEcompsocthanksitem Bangshuo Zhu, Yuekang Li, and Jingling Xue are with the School of Computer Science and Engineering, University of New South Wales, New South Wales 2052, Australia (e-mail: 
\{bang.zhu, yuekang.li, j.xue\}@unsw.edu.au).
\IEEEcompsocthanksitem Wei Song is with the School of Information and Communication Technology, Griffith University, Queensland 4111, Australia (e-mail: w.song@griffith.edu.au).
\IEEEcompsocthanksitem Yuxin Cao is with the School of Computing, National University of Singapore, Singapore 117417, Singapore (e-mail: yuxincao@u.nus.edu).
\IEEEcompsocthanksitem Yuezhong Wu is with the College of Computer and Data Science, Fuzhou University, Fuzhou Province 350025, China (e-mail: yuezhong.wu@fzu.edu.cn).
\IEEEcompsocthanksitem Zhiquan Liu is with the College of Cyber Security, Jinan University, Guangdong Province 510632, China (e-mail: zqliu@jnu.edu.cn).
\IEEEcompsocthanksitem This work has been submitted to the IEEE for possible publication. Copyright may be transferred without notice, after which this version may no longer be accessible.
}
}

\markboth{Preprint -- under review}%
{Zhu \MakeLowercase{\textit{et al.}}: Do Input-Level Defenses Transfer to Observation-Level Attacks on VideoLLMs?}


\def\eg{\emph{e.g.,}\xspace}
\def\etc{\emph{etc}\xspace}
\def\ie{\emph{i.e.,}\xspace}
\def\etal{\emph{et al.}\xspace}
\def\vs{\emph{vs.}\xspace}
\def\cf{\emph{cf.}\xspace}

\newcommand{\VideoLLMs}{\text{VideoLLMs}\xspace}
\newcommand{\VideoLLM}{\text{VideoLLM}\xspace}
\newcommand{\fName}{\textsc{DefTEval}\xspace}

\maketitle

\input{abstract}


\input{introduction}
\input{background_related_work}
\input{framework}
\input{evaluation_setup}
\input{experiment}
\input{discussion}
\input{conclusion}

\bibliographystyle{IEEEtran}
\bibliography{ref}

\vfill

\end{document}

%% file: abstract.tex
\begin{abstract}
Video Large Language Models (\VideoLLMs) are increasingly deployed in safety-critical applications such as content moderation and video analytics.
To process long videos efficiently, \VideoLLMs rely on frame sampling, token compression, and modality fusion, which together form an observation pipeline that reduces the raw video to a compact internal representation. Recent \emph{observation-level attacks} exploit this pipeline to prevent the model from perceiving harmful content, yet no defense has been explicitly designed for this threat.
We introduce \fName, a controlled evaluation framework that systematically assesses whether input-level adversarial defenses, which operate on the pixel content of already-sampled frames, can mitigate observation-level attacks. Across five \VideoLLMs, eleven representative defenses, and five attack types, we find that input-level defenses offer limited and inconsistent protection, with harmful detection rates frequently near zero. Critically, defenses fail even against attacks that embed harmful signals in \emph{every} sampled frame, indicating that the bottleneck extends beyond sampling omission to the suppression of signals that do enter the model. Token compression discards localized features, and modality fusion systematically down-weights weakened visual signals. Furthermore, defense effectiveness is dominated by model architecture rather than by the defense method itself, and detection rates vary drastically across content categories, exposing structural weaknesses in temporal reasoning. These findings demonstrate that securing \VideoLLMs requires system-level robustness mechanisms spanning sampling-aware coverage guarantees, token-level preservation of safety-relevant features, and modality-balanced fusion.
\end{abstract}

%% file: introduction.tex
\section{Introduction}
\label{sec:intro}

\IEEEPARstart{V}{ideo} Large Language Models have emerged as a foundational paradigm for multimodal AI, extending large language models to videos.
As intelligent systems increasingly operate in real-world environments, ranging from autonomous agents and assistive technologies to content moderation and video analytics, the ability to interpret videos has become critical \cite{AAAI-Yuxin, poisonVID, SecVID}. By integrating visual encoders with large language models, \VideoLLMs enable question answering over events and actions \cite{VideoQA-CVPR2024-1, VideoQA-CVPR2024-2}, video summarization \cite{VideoSum-NIPS2021, VideoSum-Survey}, and decision support for downstream applications such as retrieval, assistive perception, and safety monitoring \cite{li2024llama-vid, icsf, cpfd}.

However, the scalability of \VideoLLMs is constrained by the temporal redundancy of video streams \cite{tang2025aks, cheng2024videollama2, bluesuffix}. Exhaustively processing all frames is computationally prohibitive, and thus practical systems rely on frame sampling mechanisms \cite{tang2025aks, huang2025frag, lgs} to reduce input dimensionality. While essential for efficiency, this design creates a structural mismatch between the full video and the subset of frames actually observed by the model. As a result, the reasoning of a \VideoLLM is conditioned on partial observation rather than on the complete input.

\begin{figure*}[t]
    \centering
    \begin{subfigure}[t]{0.48\textwidth}
        \centering
        \includegraphics[width=\linewidth]{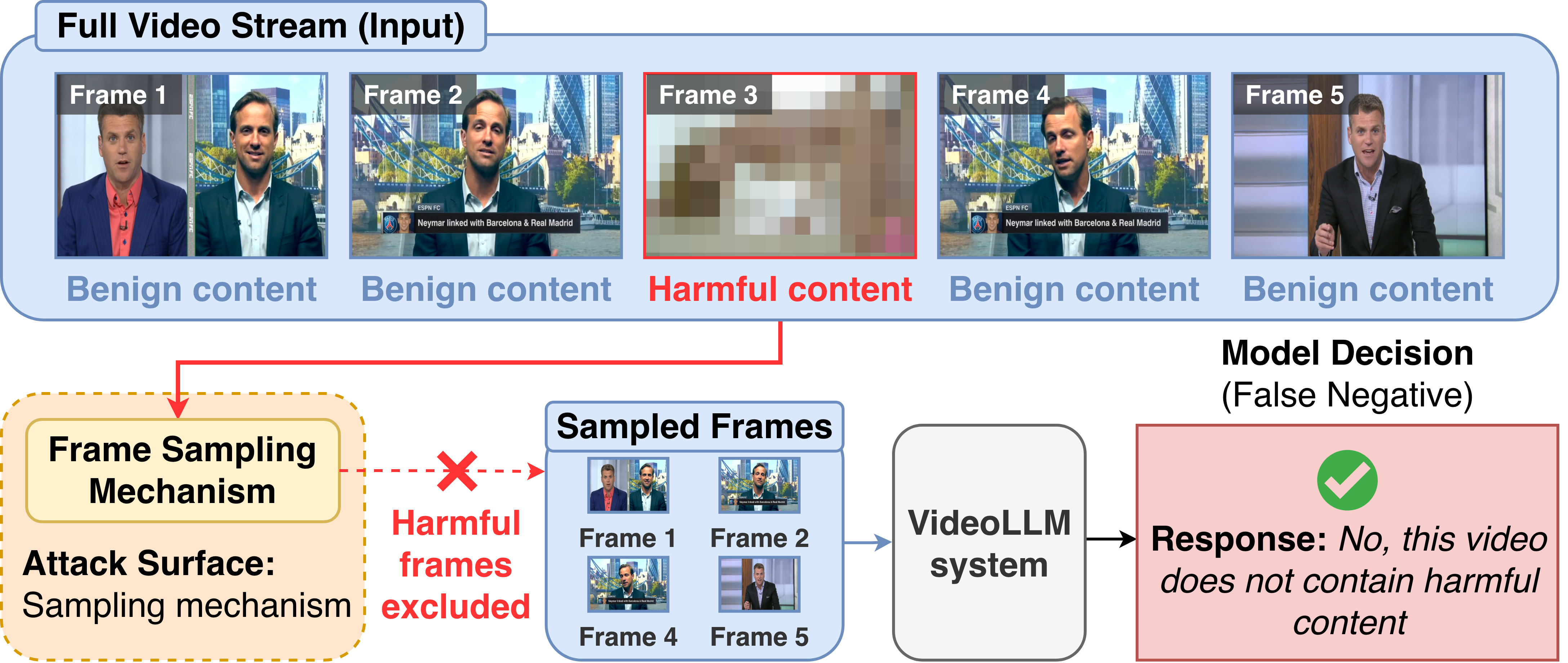}
        \caption{\textbf{Omission} (e.g., PoisonVID, FRA). The harmful segment (Frame~3) is embedded in the video but excluded during frame sampling. The \VideoLLM reasons only over benign frames and produces a false-negative safety decision.}
        \label{fig:attack_graph_1}
    \end{subfigure}
    \hfill
    \begin{subfigure}[t]{0.48\textwidth}
        \centering
        \includegraphics[width=\linewidth]{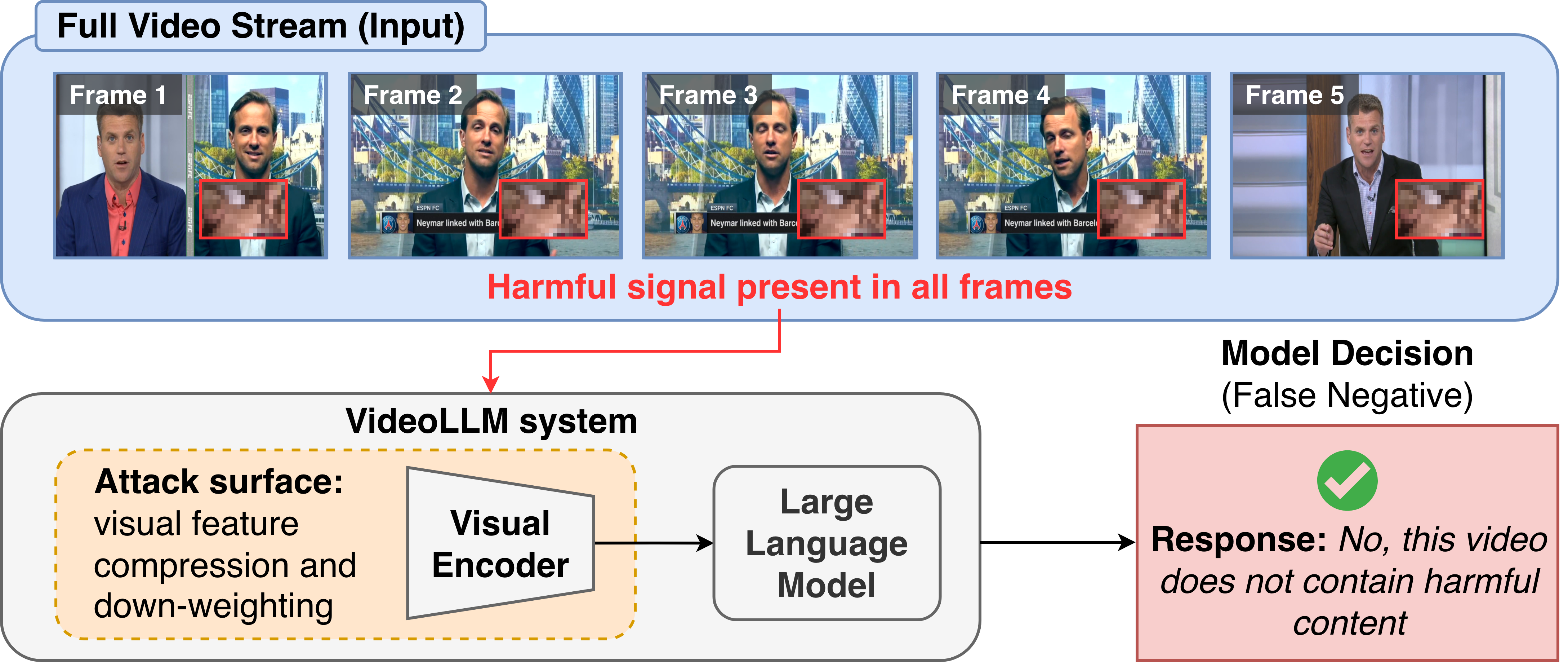}
        \caption{\textbf{Suppression} (e.g., PiP, TOA). Harmful content is present in every frame as a localized overlay, yet token compression and spatial downsampling within the visual encoder discard the fine-grained harmful features. The \VideoLLM fails to detect the harmful content despite its continuous presence in the input.}
        \label{fig:attack_graph_2}
    \end{subfigure}
    \caption{Two categories of \textbf{observation-level attacks} against \VideoLLMs. Both cause false-negative safety decisions, but through different mechanisms. In (a), harmful frames are never sampled. In (b), harmful signals enter all sampled frames but are suppressed during internal processing.
    \label{fig:attack_graph}}
    \vspace*{-2ex}
\end{figure*}

This dependence on sampled frames expands the attack surface. Instead of perturbing the observed input, an adversary may manipulate the sampling stage so that frames containing harmful or safety-relevant content are excluded from analysis. As illustrated in Fig.~\ref{fig:attack_graph}, a brief segment depicting policy-violating behavior can be embedded within an otherwise benign video yet omitted during frame selection. Consequently, \VideoLLMs deployed in content moderation, safety monitoring, or automated review settings may produce false-negative decisions despite the presence of harmful material. Such \emph{observation-level attacks} have been empirically shown to substantially undermine the reliability of \VideoLLMs \cite{AAAI-Yuxin, poisonVID}.
We define observation-level attacks as adversarial strategies that exploit the observation pipeline of \VideoLLMs, including frame sampling, token compression, and modality fusion, to prevent the model from perceiving safety-relevant content, without relying on bounded input perturbations.

\IEEEpubidadjcol

Observation-level attacks constitute a diverse threat family with multiple distinct mechanisms. Some attacks manipulate prompt-guided frame sampling so that harmful frames are never selected for inference (e.g., PoisonVID~\cite{poisonVID}). Others exploit sparse temporal sampling by embedding harmful content in short segments that are statistically unlikely to be sampled (e.g., FRA~\cite{AAAI-Yuxin}). Still others embed harmful signals persistently across all frames, as a localized spatial overlay (PiP~\cite{AAAI-Yuxin}) or a transparent blend (TOA~\cite{AAAI-Yuxin}), yet the harmful content is nonetheless suppressed during token compression, spatial downsampling, or modality fusion. These distinct attack vectors may require fundamentally different defense strategies, and the extent to which each mechanism contributes to defense failure has not been studied.

This gap motivates our central research question. To what extent can input-level adversarial defenses mitigate observation-level attacks in \VideoLLMs, and how do attack mechanisms, model architectures, and content categories determine their effectiveness or failure?
We refer to input-level defenses as methods that operate on the pixel-level content of frames that have already been sampled, without interacting with the sampling mechanism itself, including input denoising~\cite{bluesuffix}, compression and reconstruction~\cite{comdef, SecVID}, gradient smoothing~\cite{lgs}, and patch removal~\cite{pad}. Answering this question requires not only quantifying defense effectiveness but also explaining why defenses fail differently across attack types, whether model architecture matters more than the defense method, and what structural weaknesses in \VideoLLMs the varying detection rates across content categories reveal.

To address this question, we develop the \textbf{Def}ense \textbf{T}ransferability \textbf{Eval}uation Framework (\fName), a systematic evaluation framework for assessing input-level defenses under observation-level attacks in \VideoLLMs.
Using \fName, we evaluate eleven representative input-level defenses, including ComDefend \cite{comdef}, DiffPure \cite{diffp}, Image Compression CNN (ICC) \cite{icc}, Local Gradient Smoothing (LGS) \cite{lgs}, Patch-Agnostic Defense (PAD) \cite{pad}, Pixel Deflection (PIXD) \cite{pixd}, Image Quilting (Quilting) \cite{quilt}, Randomized Padding (RAND) \cite{rand}, Image Super-Resolution (SuperRes) \cite{superres}, Total Variation Minimization (TVM) \cite{tvm}, and VideoPure \cite{videopure}, against five representative \VideoLLMs (LLaVA-Video-7B-Qwen2, LLaVA-NeXT-Video-7B-DPO, LLaVA-Video-32B-Qwen, ShareGPT4Video-8B, and VideoLLaMA3) under five observation-level attacks. These attacks comprise three omission attacks, namely PoisonVID instantiated with the AKS and FRAG samplers \cite{poisonVID} and FRA \cite{AAAI-Yuxin}, and two suppression attacks, namely PiP and TOA \cite{AAAI-Yuxin}.

Our evaluation reveals four key findings.
\textbf{First}, input-level defenses provide limited and inconsistent protection. Across 825 evaluation cells, the harmful detection rate (HDR) is in the single digits or at zero for the overwhelming majority of configurations, and a four-way variance decomposition shows that \emph{which} defense is applied is the least consequential of the four experimental factors.
\textbf{Second}, this failure extends beyond sampling omission. Even when harmful signals are present in every sampled frame, defenses still fail broadly. We trace this failure to the model's own visual pathway and verify the attribution by ablating the token-compression stage, confirming that compression is what discards the harmful evidence in these cases.
\textbf{Third}, defense effectiveness is dominated by model architecture rather than by the defense method, with a subset of tested models accounting for the majority of defensive gains.
\textbf{Fourth}, detection rates vary drastically across content categories. Pornographic content, which relies on static visual cues, is detected far more often than violence or crime, which require temporal reasoning. A video-native defense does not close this gap, because sparse frame sampling strips the input of the frame-to-frame continuity such defenses are built on, disabling them before they run.


In summary, we make the following contributions:
\begin{itemize}
    \item \textbf{Controlled evaluation framework.} We introduce the \textbf{Def}ense \textbf{T}ransferability \textbf{Eval}uation Framework (\fName), the first controlled framework for assessing whether input-level adversarial defenses transfer to observation-level attacks in \VideoLLMs. \fName is a four-way full factorial design over five \VideoLLMs, eleven defenses spanning five families, five attack types, and three harmful-content categories, yielding 825 evaluation cells with the sampling budget, prompt, decoding policy, and response-to-label protocol held fixed throughout.

    \item \textbf{A mechanism taxonomy for observation-level attacks.} We formalize observation-level attacks as two distinct mechanisms. Under \emph{omission}, harmful frames are never selected for inference. Under \emph{suppression}, they are selected but do not survive the model's visual pathway. Prior work treats these as a single phenomenon. Separating them is what makes the failure of input-level defenses a structural prediction rather than an empirical observation, since a defense applied strictly after sampling can address neither.

    \item \textbf{Quantitative evidence.} Through extensive experiments, we provide the first comprehensive quantitative evidence that input-level defenses offer limited and inconsistent mitigation against observation-level attacks, with harmful detection rates frequently near zero.

    \item \textbf{Verified failure mechanisms.} We test proposed attributions of defensive breakdown directly, rather than inferring them from aggregate outcomes. Ablating the token-compression stage isolates the point in the visual pathway at which persistently present harmful signals are lost, and probing the alignment between frame embeddings and harmful-concept descriptions on both sides of each defense establishes whether a defense changes what the vision encoder recovers from the input at all. We further show that a video-native defense fares no better than its image counterparts, because sparse frame sampling removes the frame-to-frame continuity such defenses are built on.
\end{itemize}

%% file: background_related_work.tex
\section{Background and Related Work}
\label{sec:background_and_related_work}

\subsection{\VideoLLMs} 
Video Large Language Models (\VideoLLMs) extend language models from static text or image inputs to temporally structured visual streams by conditioning generation on both video content and user prompts \cite{liu2025nvila,chen2024InternVL2.5,jin2024chat-univi}. By integrating visual perception with language reasoning, \VideoLLMs enable unified video understanding and generation within a single autoregressive framework. They have rapidly emerged as a foundational paradigm for multimodal AI, supporting a wide spectrum of downstream tasks including video summarization \cite{li2023videochat,zhang2025videollama3,lin2024video-llava}, captioning \cite{yang2023vid2seq,chen2024sharegpt4video}, question answering \cite{zhang2024simple,liu2025nvila}, video grounding \cite{wang2025GroundedVideoLLM,qian2024streaming,yu2023sevila}, and long-form video comprehension \cite{weng2024longvlm,cheng2024focuschat,wang2025seal}. In these applications, the model receives a video-query pair and is expected to generate responses that are semantically consistent with and grounded in the visual evidence contained in the video \cite{li2024llava-onevision}.


\subsection{Frame Sampling Strategies}
Frame sampling methods can be categorized into three strategies: uniform sampling, semantic similarity sampling, and prompt-guided sampling \cite{poisonVID}.

Uniform frame sampling selects frames at fixed temporal intervals, typically ensuring inclusion of the first and last frames \cite{zhang2024llavanextvideo}. This strategy is widely adopted in systems such as Apollo \cite{zohar2025apollo} and various LLaVA- and LLaMA-based models \cite{zhang2024llavanextvideo,lin2024video-llava,cheng2024videollama2}. Its simplicity and low overhead make it attractive in practice. However, by fixing sampling positions regardless of content distribution, uniform sampling often overlooks informative segments, especially in longer videos where salient events are sparsely distributed \cite{zohar2025apollo,hu2025mllm}.

To improve content coverage, semantic similarity-based sampling strategies perform denser initial sampling and then remove redundant frames based on visual similarity. For example, Semantic-aware Key-frame Extraction (SKE) \cite{chen2024sharegpt4video} computes frame-level embeddings (e.g., using CLIP~\cite{radford2021clip}) and measures pairwise cosine similarity to eliminate highly similar frames. By maximizing semantic diversity among retained frames, semantic similarity-based sampling aims to preserve representative visual content. However, because selection is agnostic to the user query, it may still retain frames irrelevant to the prompt. Moreover, similar to uniform frame sampling, some implementations always include boundary frames, which are frequently uninformative (e.g., blank or static scenes) \cite{tang2025aks}.

More recently, prompt-guided sampling methods incorporate query relevance into the selection process. These approaches first sample candidate frames and then compute frame-prompt relevance scores using either a lightweight vision-language model (e.g., BLIP~\cite{li2023blip}) or other scoring mechanisms. Frames with higher relevance are prioritized for downstream encoding, improving localization and task performance. Representative methods include Differential Keyframe Selection (DKS) \cite{cheng2025vilamp}, Adaptive Keyframe Sampling (AKS) \cite{tang2025aks}, and Frame Selection Augmented Generation (FRAG) \cite{huang2025frag}. DKS further accounts for local feature redundancy, AKS introduces coverage terms to encourage diversity, and FRAG emphasizes relevance ranking. While prompt-guided approaches generally improve task alignment, they may discard context that is weakly correlated with the prompt yet still semantically important. In addition, certain methods (e.g., FRAG) rely on full-scale \VideoLLMs for relevance estimation, increasing computational overhead. 


\subsection{Observation-Level Attacks} 
Existing research on \VideoLLMs has largely focused on enhancing model capability and task performance, with comparatively limited attention to security risks arising from their input processing pipelines. Recent work has begun to explore such vulnerabilities. In particular, \cite{AAAI-Yuxin} identifies structural weaknesses in current architectures, including sparse temporal sampling, token under-sampling, and modality fusion imbalance, that enable uniform frame sampling- and semantic similarity sampling-based \VideoLLMs to miss harmful content even when it is deliberately embedded in the video.
PoisonVID \cite{poisonVID} examines the security implications of more advanced sampling strategies. Rather than targeting uniform or similarity-based sampling alone, it investigates whether prompt-guided methods such as DKS \cite{cheng2025vilamp}, AKS \cite{tang2025aks}, and FRAG \cite{huang2025frag} remain susceptible to manipulation.

\subsection{Adversarial Defenses}
\label{subsec:adversarial_defenses}

Existing adversarial defenses can be broadly grouped into adversarial training, input reconstruction, gradient smoothing, and patch removal approaches.

\subsubsection{Adversarial Training} These methods \cite{FARE, simclip} enhance resilience by improving feature stability under adversarial perturbations. In \VideoLLMs, adversarial training-based defenses commonly strengthen robustness by adversarially fine-tuning the vision encoder with augmented datasets, aiming to improve the stability of visual representations against bounded perturbations while leaving the language module unchanged. Representative works such as FARE \cite{FARE} and SimCLIP \cite{simclip} adversarially fine-tune CLIP \cite{radford2021clip}, improving the stability of visual representations under perturbation-based threat models.

\subsubsection{Input Reconstruction} These defenses \cite{comdef, diffp, cider, icc, tvm, pixd, rand, quilt, superres} mitigate adversarial effects at inference time by suppressing or removing perturbations from the input signal. Reconstruction frameworks attempt to purify inputs by reconstructing visually faithful images from adversarial samples while filtering out adversarial noise, using techniques such as a compression-reconstruction network \cite{comdef} or generating a close facsimile with an assembly of small, clean patches \cite{quilt}. Denoise-only techniques such as ICC \cite{icc}, PIXD \cite{pixd}, SuperRes \cite{superres}, and TVM \cite{tvm} perform no reconstruction, but instead attenuate adversarially perturbed regions of an image relative to its salient regions. Diffusion-based purification methods \cite{bluesuffix} similarly denoise corrupted inputs through iterative stochastic processes. Some systems, such as CIDER \cite{cider}, incorporate an explicit detection stage prior to reconstruction to prevent unnecessary degradation of benign inputs. Despite differences in implementation, these approaches share a common premise: adversarial influence is assumed to appear as additive or structured perturbations on an otherwise fully observable input.

\subsubsection{Gradient Smoothing} These methods mitigate adversarial perturbations by suppressing abnormal, localized gradient responses in feature representations. These methods are motivated by the observation that patch-based attacks often introduce spatially concentrated perturbations that distort activation patterns. Local Gradient Smoothing (LGS) \cite{lgs} exemplifies this approach by identifying regions with abnormal gradient magnitude and applying localized smoothing to attenuate adversarial influence. Such defenses assume that malicious perturbations are embedded within an otherwise fully observable input and manifest as spatially localized anomalies.

\subsubsection{Patch Removal} This category of defense deals specifically with adversarial patch attacks \cite{patch1, patch2}, where the adversarial noise is confined to a small, localized patch of the target image. As opposed to traditional adversarial perturbations, where noise must be present in the entire image, the bounded area of an adversarial patch makes it a more realistic threat model in physical vision applications, as an adversarial patch can be printed out and used in the physical world \cite{patch1}. Defenses against these attacks typically identify an adversarial patch then remove it with techniques like masking \cite{patchcleanser, cbm, sac} or regularizing local region gradients \cite{lgs}.

\subsubsection{Temporal Video Defenses} A small number of defenses operate on video rather than on individual frames, exploiting temporal structure that per-frame methods discard. VideoPure \cite{videopure} exemplifies this class. It purifies a clip through diffusion in the latent space of a pretrained text-to-video model, denoising all frames jointly rather than independently, and guides the reverse process with optical flow estimated between adjacent frames so that the purified clip remains temporally coherent. Such defenses assume that the input is a temporally contiguous sequence whose consecutive frames are related by continuous motion, an assumption the frame sampling stage of a \VideoLLM does not preserve.

%% file: framework.tex
\section{\fName}
\label{sec:framework}

\subsection{Formalism}
\label{sec:formalism}

Let a video be denoted by $V = (x_1, x_2, \ldots, x_T)$, where $x_t$ is the $t$-th frame and $T$ is the video length.
Given a user query $q$, a \VideoLLM first applies a frame sampling policy to select a subset of frames for inference.
We model the sampling as
\begin{equation}
S = \pi(V, q), \qquad S \subseteq \{1,\ldots,T\}, \qquad |S| = K,
\label{eq:sampling}
\end{equation}
where $\pi(\cdot)$ denotes the sampling mechanism (e.g., uniform frame sampling, semantic similarity sampling, or prompt-guided sampling), and $K$ is the number of sampled frames.
Let $V_S \triangleq \{x_t \mid t \in S\}$ denote the sampled frames.
The \VideoLLM then produces an output (e.g., a textual description)
\begin{equation}
y = f(V_S, q; \theta) = h\big(g(V_S), q; \theta\big),
\label{eq:videollm}
\end{equation}
where $f(\cdot)$ denotes the end-to-end \VideoLLM parameterized by $\theta$, decomposed into a visual pathway $g(\cdot)$ (encoder, projector, and token compression) and a fused language model $h(\cdot)$ that integrates the resulting visual tokens with the query.
Equations~\eqref{eq:sampling} and \eqref{eq:videollm} make explicit that the model's prediction is conditioned on the \emph{observed subset} $V_S$, and further on the compressed representation $g(V_S)$ rather than on $V_S$ itself.

We consider \emph{observation-level adversarial attacks} that exploit this observation pipeline.
Unlike conventional adversarial examples that perturb the observed inputs by introducing bounded perturbations (e.g., $x_t \mapsto x_t+\delta_t$ with $\|\delta_t\|\leq \epsilon$), the attacker here does not rely on bounded input perturbations.
Instead, the attacker constructs an \emph{attacked video} $V'=\mathcal{A}(V)$ by inserting a harmful segment into an otherwise benign video, enabling policy-violating content to bypass \VideoLLM-based video moderation and safety monitoring \cite{AAAI-Yuxin, poisonVID}.
Formally, let $S=\pi(V,q)$ be the nominal selection set for a benign video-query pair $(V,q)$.
Under attack, the \VideoLLM receives $V'=\mathcal{A}(V)$ and selects
\begin{equation}
S' = \pi(V',q), \qquad V'=\mathcal{A}(V),
\label{eq:obs_attack}
\end{equation}
yielding the attacked output
\begin{equation}
y' = f(V'_{S'}, q; \theta).
\label{eq:attacked_output}
\end{equation}
Let $\mathcal{H}(V') \subseteq \{1,\ldots,T\}$ denote the indices of frames in $V'$ that contain harmful or safety-relevant content.
The attacker's objective is that $y'$ remain consistent with the benign response $y$ even though harmful content is present in $V'$.
Two distinct mechanisms suffice to achieve this objective, namely
\begin{equation}
\underbrace{S' \cap \mathcal{H}(V') = \emptyset}_{\text{omission}}
\quad\text{or}\quad
\underbrace{g(V'_{S'}) \approx g(V_S)}_{\text{suppression}} ,
\label{eq:two_mechanisms}
\end{equation}
where the omission condition, and in practice its relaxation $|S' \cap \mathcal{H}(V')| \ll K$, states that few or no harmful frames are selected, and the suppression condition states that harmful frames are selected but do not survive visual encoding and token compression, so the attacked and benign observations become indistinguishable to the language model.
Omission is an objective the attacker pursues rather than one it can guarantee, and the residual harmful frames left behind when it falls short are precisely what an input-level defense has left to work with.
The distinction matters for defense. Under omission the evidence is largely gone before any defense can act on it, whereas under suppression it is present in full at the defense's input and is discarded only afterwards.
Section~\ref{sec:experiment} shows that both mechanisms defeat input-level defenses, and that the second is the more consequential of the two.

\subsection{Threat Model}
\label{sec:threat}
We consider an adversary who aims to bypass \VideoLLM-based video moderation or safety monitoring by inducing \emph{false-negative} outputs, while leaving the underlying model parameters unchanged.
The adversary can construct an attacked video $V'=\mathcal{A}(V)$ by inserting, replacing, or temporally relocating a short harmful segment within an otherwise benign video, or by overlaying one persistently across every frame.
We assume the attacker does not modify the \VideoLLM parameters $\theta$ and does not rely on bounded input perturbations on the observed frames.

The adversary's leverage is at the observation stage. By manipulating the input video (and, for prompt-guided sampling, potentially the query or relevance cues), the attacker aims to satisfy either condition in Equation~\eqref{eq:two_mechanisms}.
Unless stated otherwise, we assume the attacker knows the number of frames ($K$) sampled by the \VideoLLM, but does not require access to the model's internals.
Attacks that rely on suppression rather than omission do not require even this knowledge, and are fully black-box.

\subsection{Controlled Evaluation Design}
\label{sec:design}

\fName{} is a controlled factorial design over the composition of a defense with an attacked observation pipeline.
Prior work establishes that observation-level attacks succeed against undefended \VideoLLMs \cite{AAAI-Yuxin, poisonVID}, that is, that $y'$ in Equation~\eqref{eq:attacked_output} is benign-consistent when $V'$ is attacked.
\fName{} asks the question that follows from this result. Given an input-level defense $d(\cdot)$ applied to the frames the pipeline has already selected, does
\begin{equation}
y'_d = f\big(d(V'_{S'}),\, q;\, \theta\big)
\label{eq:defended}
\end{equation}
recover the harmful evidence, and which properties of the configuration determine whether it does?
Equation~\eqref{eq:defended} makes the scope of an input-level defense explicit. Because $d$ is applied strictly after $\pi$, it can act only on $V'_{S'}$.
It can never restore the frames $\mathcal{H}(V') \setminus S'$ that sampling discarded, and it operates entirely upstream of $g$, so it cannot influence which of its outputs survive token compression or how they are weighted during fusion.
This scoping is what makes the failure of input-level defenses a structural prediction rather than an empirical accident.

\smallskip
\noindent
\textbf{Factors and levels.}
We treat the evaluation as a four-way full factorial design over the factors that a deployer can actually choose or encounter, namely the \emph{model architecture} (5 levels), the \emph{defense} (11 levels, spanning five families), the \emph{attack} (5 levels, spanning the two mechanisms of Equation~\eqref{eq:two_mechanisms}), and the \emph{harmful-content category} (3 levels).
Crossing all four yields $5 \times 11 \times 5 \times 3 = 825$ evaluation cells, each estimated from 100 attacked videos.
Full crossing is what distinguishes \fName{} from a defense benchmark. It is the only design under which the contribution of the defense can be separated from that of the model, the attack, and the content, rather than confounded with them.

\smallskip
\noindent
\textbf{Controlled quantities.}
Every factor not under study is held fixed across all 825 cells.
All attacks are constructed against the same sampling budget $K$ and the same inference configuration.
Each model is queried with a single fixed prompt template per content category, and is decoded deterministically (\texttt{do\_sample} disabled), so that no cell varies through generation randomness.
Model responses are mapped to binary decisions by one fixed protocol, described in Section~\ref{sec:setup}.
Each attacked video carries exactly one harmful segment.

\smallskip
\noindent
\textbf{The undefended baseline is zero by construction.}
Every attacked video entering the study has already been verified to defeat the target model without any defense applied.
The measured quantity, HDR (Equation~\eqref{eq:hdr}), is therefore conditioned on the attack having succeeded. It reports the fraction of \emph{already-failed} cases that a defense recovers, and the no-defense baseline is identically $0\%$ in every cell.
This removes the principal confound in defense evaluation, namely that a defense may appear effective merely because the underlying attack was weak, and means any non-zero HDR is attributable to the defense alone.

\smallskip
\noindent
\textbf{Analysis plan.}
The design supports two complementary analyses, fixed in advance.
First, the distribution of HDR over all cells characterizes how often \emph{any} configuration achieves meaningful recovery, independently of which configuration it is.
Second, a four-way ANOVA with $\eta^2$ effect sizes apportions the variance in HDR among the four factors, answering which of them governs the outcome. All interaction terms are pooled into the residual, so the reported $\eta^2$ values are conservative lower bounds on each factor's influence.
Together these analyses answer the two halves of our research question, namely how well input-level defenses transfer and what determines whether they transfer at all.

\smallskip
\noindent
\textbf{Relation to prior work.}
The attacks we evaluate were introduced by \cite{AAAI-Yuxin} and \cite{poisonVID}, which measure \emph{attack success} against undefended models and attribute it to sampling sparsity and prompt-guided relevance manipulation, respectively.
Neither work evaluates a defense.
\fName{} addresses the complementary and previously open question of whether the existing input-level defense literature transfers to this threat model. It is the factorial design above, rather than any individual measurement, that makes the question answerable. A single defense and model pair cannot distinguish a weak defense from a model that no defense can help, whereas the variance decomposition over 825 crossed cells can.

%% file: evaluation_setup.tex
\section{Evaluation Setup}
\label{sec:setup}

\subsection{\VideoLLMs}
We evaluate five advanced \VideoLLMs: LLaVA-Video-7B-Qwen2 (L-7B), LLaVA-NeXT-Video-7B-DPO (L-7B-DPO), LLaVA-Video-32B-Qwen (L-32B) \cite{llava-video}, ShareGPT4Video-8B (SG4V) \cite{chen2024sharegpt4video}, and VideoLLaMA3 (VL3) \cite{zhang2025videollama3}.
All models are tested under deterministic generation settings, with \texttt{do\_sample} set to \texttt{False}, ensuring that model behavior is not affected by sampling variance.

\subsection{Attacks}
\input{attacks_table}
Table~\ref{tab:attacks} summarizes the five observation-level attacks we evaluate, namely PoisonVID instantiated over two prompt-guided samplers \cite{poisonVID}, and the Frame Replacement (FRA), Picture-in-Picture (PiP), and Transparent Overlay (TOA) attacks of \cite{AAAI-Yuxin}.
As formalized in Equation~\eqref{eq:two_mechanisms}, omission attacks aim to keep harmful frames out of the sampled subset. When omission is incomplete, residual harmful frames survive into the observation. These residual cases carry interpretive weight. They are the only PoisonVID and FRA instances in which an input-level defense has any harmful evidence to act on at all, and they account for the non-zero HDR observed under those attacks in Table~\ref{tab:defenses_combined}. Suppression attacks place harmful content in every sampled frame by construction and rely on it being discarded somewhere along the visual pathway. For PiP and TOA, the injected content is deliberately kept perceptible. The overlay region is large enough to be read at a glance and loops for the full duration, and the blending coefficient $\alpha$ is chosen so that the harmful clip remains clearly visible. This is what makes their failure mode consequential. A human moderator reviewing the same video would see the violation immediately, while the \VideoLLM does not. All attacks are constructed against the same sampling budget, $K=32$.

\subsection{Defenses}
\input{defenses_table}
We evaluate eleven defenses, summarized in Table~\ref{tab:defenses_taxonomy}.
Selection follows two criteria.
First, a defense must be applicable at inference time without retraining or modifying the \VideoLLM.
This excludes adversarial-training approaches such as FARE \cite{FARE} and SimCLIP \cite{simclip}, which adversarially fine-tune the vision encoder and therefore alter $\theta$, violating the fixed-model assumption of our threat model (Section~\ref{sec:threat}).
Second, a defense must have a public, reproducible implementation.
Ten of the eleven are image defenses spanning the five families identified in Table~\ref{tab:defenses_taxonomy}. The eleventh, VideoPure \cite{videopure}, is a video defense included to test whether operating on the temporal dimension changes the outcome.

The ten image defenses differ in mechanism but share a single premise, namely that the adversarial signal is a bounded, structured perturbation superimposed on an input the model can otherwise fully observe.
\emph{Reconstruction and purification} methods re-synthesize the frame from a clean prior, using a learned compression and reconstruction network (ComDefend), a forward-then-reverse diffusion process (DiffPure), a database of clean image patches (Quilting), or a super-resolution network that remaps off-manifold samples back onto the natural image manifold (SuperRes).
\emph{Compression and denoising} methods attenuate perturbed regions relative to salient ones through lossy re-encoding (ICC), pixel redistribution followed by wavelet denoising (PIXD), or total variation minimization (TVM).
\emph{Input randomization} (RAND) destroys the spatial structure of perturbations by random resizing and padding.
\emph{Patch localization and removal} methods target the bounded-patch threat model, either regularizing regions of abnormal gradient magnitude (LGS) or localizing patches via semantic independence and spatial heterogeneity before suppressing them (PAD). Each image defense is applied independently to every sampled frame, and the purified frame set is passed unchanged to the downstream \VideoLLM pipeline.

We include VideoPure to test whether input-level defenses fail simply because they are applied frame by frame and therefore discard the temporal dimension. If that were the whole account, a defense that models the clip jointly should recover harmful content the image defenses miss. In our setting, VideoPure operates on the full 32-frame sampled clip at once, purifying with a shallow inversion timestep and flow guidance from a RAFT estimator \cite{raft}, then resampling to the source resolution before inference.

\subsection{Data and Metric}
We follow the threat model and attack construction protocols in prior work \cite{AAAI-Yuxin, poisonVID}. Specifically, we instantiate harmful content using three policy-violating categories, namely explicit sexual content, crime-related content, and violence, to reflect common safety-critical moderation scenarios. For each attack, we construct 100 successfully attacked videos by inserting a harmful clip into an otherwise benign source video. Unless stated otherwise, each attacked example contains a single harmful segment and has already been verified as successfully attacked, and all attacks are evaluated under the same $K$ and inference configuration to enable controlled comparisons.

Our primary metric measures how often a \VideoLLM correctly flags the presence of harmful content after applying a defense.
Let $\mathcal{D}_\text{harm}$ denote the set of videos that contain harmful content (ground truth $z=1$).
Given a defense method $d(\cdot)$, we run the model on each input and map the \VideoLLM's response to a binary decision $\hat{z}\in\{0,1\}$, where $\hat{z}=1$ indicates that the model identifies the queried harmful content and $\hat{z}=0$ indicates it does not. Let $\mathcal{D}_\text{test} \subseteq \mathcal{D}_\text{harm}$ denote the subset of harmful videos that the undefended model fails to detect, that is, those with $z=1$ and $\hat{z}=0$ when no defense is applied. We report the \emph{Harmful Detection Rate (HDR)}
\begin{equation}
\label{eq:hdr}
\mathrm{HDR}(d)=\frac{1}{|\mathcal{D}_\text{test}|}\sum_{(V,q)\in \mathcal{D}_\text{test}} \mathbb{I}\!\left[\hat{z}\big(f(d(V),q)\big)=1\right],
\end{equation}
which is the fraction of previously undetected harmful videos that are correctly detected once defense $d$ is applied (higher is better).

For the \VideoLLM's response, a clear ``yes'' answer such as ``Yes, the video appears to contain explicit content that is typically associated with adult entertainment.'' is marked as an identification, and a clear ``no'' answer such as ``No, the video does not contain pornographic content[...]'' is marked as a negative identification. Responses like ``I'm sorry, but I can't assist with that.'' are treated as ``yes'' answers, as such refusals are assumed to be triggered by the model's internal safety alignment. Responses that answer ``yes'' but explicitly refer to a subject other than the harmful content are considered false positives and treated as ``no'' answers.

%% file: attacks_table.tex
\begin{table*}[!t]
\centering
\setlength{\tabcolsep}{3pt}
\footnotesize
\begin{tabular}{@{}
  >{\raggedright\arraybackslash}p{0.175\textwidth}
  >{\raggedright\arraybackslash}p{0.105\textwidth}
  >{\raggedright\arraybackslash}p{0.190\textwidth}
  >{\raggedright\arraybackslash}p{0.185\textwidth}
  >{\raggedright\arraybackslash}p{0.235\textwidth}
  @{}}
\toprule
Attacks & Mechanism & Injection & Stage exploited & Harmful signal in sampled frames \\
\midrule

PoisonVID (AKS) \cite{poisonVID}\newline
PoisonVID (FRAG) \cite{poisonVID}\newline
FRA \cite{AAAI-Yuxin}
  & Omission
  & A harmful clip is inserted into, or substituted for a segment of, an otherwise benign video
  & The frame selection policy $\pi$, via its relevance scoring (PoisonVID) or its temporal sparsity (FRA)
  & Ideally none; residual harmful frames may survive when omission is incomplete \\
\addlinespace[4pt]
\midrule
\addlinespace[2pt]

PiP \cite{AAAI-Yuxin}\newline
TOA \cite{AAAI-Yuxin}
  & Suppression
  & A harmful clip is composited into every frame, as a fixed-position overlay (PiP) or by alpha-blending at opacity $\alpha$ (TOA)
  & The visual pathway $g$: token compression and spatial downsampling (PiP), or modality fusion imbalance (TOA)
  & Every frame, by construction; spatially localized (PiP) or attenuated (TOA) \\

\bottomrule
\end{tabular}
\caption{The five observation-level attacks evaluated in \fName, grouped by the mechanism of
Equation~\eqref{eq:two_mechanisms} they realize.}
\label{tab:attacks}
\end{table*}

%% file: defenses_table.tex
\begin{table*}[!t]
\centering
\setlength{\tabcolsep}{3pt}
\footnotesize
\begin{tabular}{@{}
  >{\raggedright\arraybackslash}p{0.14\textwidth}
  >{\raggedright\arraybackslash}p{0.20\textwidth}
  >{\raggedright\arraybackslash}p{0.60\textwidth}
  @{}}
\toprule
Defense & Family & Assumed adversary \\
\midrule

ComDefend \cite{comdef}
  & \multirow{4}{*}{\parbox{0.20\textwidth}{\raggedright Reconstruction / purification}}
  & \multirow{4}{*}{\parbox{0.60\textwidth}{\raggedright An additive perturbation that displaces an otherwise fully observable frame off the natural image manifold, and can be undone by re-synthesizing the frame from a clean prior.}} \\[2pt]
DiffPure \cite{diffp}    & & \\[2pt]
Quilting \cite{quilt}    & & \\[2pt]
SuperRes \cite{superres} & & \\
\midrule

ICC \cite{icc}
  & \multirow{3}{*}{\parbox{0.20\textwidth}{\raggedright Compression / denoising}}
  & \multirow{3}{*}{\parbox{0.60\textwidth}{\raggedright A low-amplitude, high-frequency perturbation concentrated away from salient regions, removable by attenuating fine detail without reconstruction.}} \\[2pt]
PIXD \cite{pixd} & & \\[2pt]
TVM \cite{tvm}   & & \\
\midrule

RAND \cite{rand}
  & Input randomization
  & A perturbation whose spatial structure is brittle: it survives only at the scale and alignment at which it was optimized. \\
\midrule

LGS \cite{lgs}
  & \multirow{2}{*}{\parbox{0.20\textwidth}{\raggedright Patch localization \& removal}}
  & \multirow{2}{*}{\parbox{0.60\textwidth}{\raggedright A bounded adversarial patch, localizable by its abnormal local gradients or by its semantic independence from the surrounding scene.}} \\[2pt]
PAD \cite{pad} & & \\
\midrule

VideoPure \cite{videopure}
  & Video / temporal
  & A perturbation spanning the clip that is temporally consistent, and can be suppressed by denoising frames jointly under a motion constraint. \\

\bottomrule
\end{tabular}
\caption{The eleven defenses evaluated in \fName, grouped by family and by the adversary each family assumes.}
\label{tab:defenses_taxonomy}
\end{table*}

%% file: experiment.tex
\section{Results and Analysis}
\label{sec:experiment}

Table~\ref{tab:defenses_combined} presents the harmful detection rate (HDR, Equation~\eqref{eq:hdr}) for each combination of eleven defenses, five \VideoLLMs, five observation-level attacks, and three content categories, yielding 825 evaluation cells.
We organize our analysis around four findings that collectively characterize the transferability of input-level defenses to observation-level threats and directly address the research question posed in Section~\ref{sec:intro}.
Each finding presents its empirical evidence together with the mechanism that explains it, and, where we can test that mechanism directly, the ablation that does so.

\input{table1}

\subsection{Finding~1: Input-Level Defenses Provide Limited and Inconsistent Protection}
\label{sec:finding1}

Table~\ref{tab:defenses_combined} shows that the vast majority of configurations report HDR values in the single digits or at zero, with only a minority attaining moderate or high HDR, reinforcing our conclusion that input-level defenses provide limited and inconsistent protection against observation-level attacks.

\smallskip
\noindent
\textbf{Prompt-guided sampling attacks are particularly resilient.} Under PoisonVID (both AKS and FRAG), defenses are least effective. For L-7B-DPO, both PoisonVID variants yield 0\% HDR across \emph{all} categories and \emph{all} eleven defenses. Even on other models, PoisonVID detection rates typically remain in the single digits. DiffPure on L-7B achieves only 3 to 5\% under AKS and 5 to 15\% under FRAG, and PAD shows similar magnitudes. This outcome is consistent with the core threat model. When sampling manipulation excludes harmful frames from inference, defenses that operate on the already-sampled frames have no harmful evidence to recover.

\input{anova_fourway}

\smallskip
\noindent
\textbf{Choice of defense is inconsequential.} As shown in Table~\ref{tab:anova_fourway}, of the four experimental factors, the defense is the least important determinant of HDR. Model architecture ($\eta^2 =$ 25.4\%), harmful-content category (9.4\%), and attack type (7.9\%) each explain more variance than which of the eleven defenses is applied (3.1\%). Statistically, the choice of defense is thus the least consequential choice in the entire evaluation. These results confirm that input-level defenses offer, at best, narrow and attack-specific improvements rather than general protection against observation-level threats.

\smallskip
\noindent
\textbf{Why omission attacks resist defense.} For PoisonVID, the adversary directly targets the relevance scoring mechanism in prompt-guided sampling, suppressing the selection probability of harmful frames so that they are systematically excluded from the sampled subset. Input-level defenses, which operate on frames \emph{after} sampling, cannot recover evidence that was never selected. Similarly, FRA exploits the sparsity of uniform temporal sampling. By localizing the harmful segment in a short interval, it ensures that the probability of any sampled frame falling within the injected region is low. This constitutes a fundamental threat-model mismatch, since input-level defenses are designed to purify observed content, not to ensure coverage of unobserved content.

\input{siglip}

\smallskip
\noindent
\textbf{Defenses do not restore harmful semantics.} To test whether defenses recover safety-relevant content rather than merely altering pixels, we embed all $K{=}32$ sampled frames with the SigLIP-so400m-384 tower \cite{zhai2023siglip} and measure their cosine similarity to harmful-concept prompts for the video's category, before and after each defense. The two attack mechanisms require different measures. Under omission, the harmful clip occupies roughly one sampled frame, so we track that frame's prominence within its own clip (max $-$ median), which is unaffected by uniform shifts in the embedding distribution. Under suppression, the clip is composited into every frame and no such contrast exists, so we instead isolate the concept-specific part of the shift, defined as the change under the video's own harmful concept minus the change under the two non-matching concepts, which cancels global drift. Across both mechanisms (Table~\ref{tab:siglip}), harmful alignment never meaningfully increases, with the only positive entries being $+1.3\%$ (ComDefend, omission) and $+0.1\%$ (LGS, suppression). Input-level defenses therefore have no mechanism to restore safety-relevant signal. They can only remove perturbation, which is the wrong operation against attacks that hide harmful content in unperturbed frames.

\subsection{Finding~2: Defenses Fail Even When Harmful Signals Persist in Every Frame}
\label{sec:finding2}

\input{ablation_table2}

A critical test of input-level defenses is their performance under \emph{suppression} attacks, such as PiP (spatially localized overlay) and TOA (transparent alpha-blending), where harmful content is present in \emph{every} sampled frame. If defense failure were attributable solely to sampling omission, these attacks should be substantially easier to defend against. This is not the case. Suppression attacks yield lower detection rates overall. Pooled over all defenses, models, and categories, mean HDR is 4.9\% for the omission attacks against 3.0\% for the suppression attacks.

\smallskip
\noindent
\textbf{TOA defeats all defenses across nearly all configurations.}
Under TOA, HDR remains below 5\% across most model and defense combinations. For L-7B and L-7B-DPO, nearly all cells report 0 to 2\%. Even the best-performing defense and model pair (PAD on L-32B) reaches only 23\% on Porn and 11\% on Crime, while all other TOA cells for L-32B stay below 10\%.

\smallskip
\noindent
\textbf{PiP shows sporadic recovery, but only for specific model and category pairs.} Under PiP, results are similarly suppressed for most models, with the notable exception of L-32B on Porn. Yet even in these cases, Crime and Violence categories under PiP remain substantially lower, indicating that recovery is highly category-dependent rather than a general defense capability.

\smallskip
These results demonstrate that the failure of input-level defenses extends beyond sampling omission. We hypothesize that token compression and spatial downsampling, applied to fit long videos within the context window, selectively discard visual tokens from localized or low-contrast regions, stripping away the harmful features even when they are spatially present. Furthermore, during multimodal integration, the modality fusion mechanism systematically down-weights attenuated visual signals relative to the language pathway. As a result, the model never effectively processes the safety-relevant evidence, even though it physically enters the input.

\smallskip
\noindent
\textbf{Removing token compression restores detection.}
We test this hypothesis by ablating the compression stage of the tested models. We run the ablation on DiffPure and PAD, the two strongest defenses against PiP and TOA. Every \VideoLLM compresses its visual tokens, but through a different mechanism, so the ablation differs per architecture. For the LLaVA-NeXT-Video models we set \texttt{mm\_spatial\_pool\_stride} from 2 to 1, removing the $2{\times}2$ pool over each frame's patch grid. SG4V has no spatial pool and instead tiles all 32 frames into a single montage, so we widen \texttt{image\_grid\_pinpoints} from $2{\times}2$ to $4{\times}4$. For VL3 we disable \texttt{use\_token\_compression} and set \texttt{video\_merge\_size} to 1.

Removing compression raises pooled HDR from 4.8\% to 11.0\%, and the gain concentrates in exactly the models the hypothesis predicts (Table~\ref{tab:ablation_compression}). L-32B gains $+17.0$ points on Crime and $+14.8$ on Porn, and VL3 gains $+28.0$ on Porn. The two models that detect almost nothing under any defense, L-7B and SG4V, barely move ($+0.0$ to $+4.2$), because where the underlying representation carries no usable harmful signal, restoring the token budget has no effect.

This finding is non-trivial. It demonstrates that the problem is not merely a threat-model mismatch, but an \emph{architectural fragility} in how \VideoLLMs process and integrate visual information.

\subsection{Finding~3: Defense Effectiveness Is Dominated by Model Architecture}
\label{sec:finding3}

As shown in Table~\ref{tab:anova_fourway}, model architecture explains roughly eight times more variance in HDR than the choice of defense ($\eta^2 =$ 25.4\% against 3.1\%), with detection capability concentrated in two models. Averaged over all defenses, attacks, and categories, L-32B and VL3 reach 8.2\% and 7.9\% HDR, against 2.4\% for L-7B, 2.0\% for SG4V, and 0.1\% for L-7B-DPO. This gap between the two groups holds under every one of the eleven defenses, and the stronger group accounts for 78\% of all harmful detections.

\smallskip
\noindent
\textbf{Consistently low-HDR models.}
L-7B-DPO yields HDR near 0\% under virtually all 120 defense, attack, and category configurations.
SG4V shows similarly negligible rates, at 0 to 2\% under PoisonVID and PiP, with only isolated non-zero values under PAD and DiffPure for specific categories.
Notably, switching the defense method, from simple compression (ICC) to diffusion-based purification (DiffPure) to patch removal (PAD), produces no meaningful improvement on either model.

\smallskip
\noindent
\textbf{Comparatively higher-HDR models.}
L-32B shows moderate recovery under specific conditions. Under PoisonVID (FRAG) on Porn, it reaches 41\% (PAD), 37\% (DiffPure), and 32\% (ICC).
VL3 exhibits elevated Porn detection under several attacks, reaching 34\% (ICC), 31\% (ComDefend), and 27\% (PIXD) under FRA.
However, even on these models, the majority of cells remain below 15\%.

\smallskip
This pattern indicates that model-internal factors, including visual representation quality, token retention policy, and safety alignment strategy, constitute the primary bottleneck for defense effectiveness.
These factors jointly determine whether a defense can extract residual harmful signals from the processed input. When they are unfavorable, no input-level defense can compensate.
From a deployment perspective, the choice of model architecture may be more consequential for robustness than the choice of defense method.

\subsection{Finding~4: Content-Category Disparity Exposes Temporal Reasoning Weakness}
\label{sec:finding4}

The highest HDR values in Table~\ref{tab:defenses_combined} are overwhelmingly concentrated in the Porn category, which contains the ten highest HDR values in the entire table. Under FRA and PiP, the Porn advantage persists, while Crime and Violence detection rates rarely exceed 15\% under any combination of defense, model, and attack.

This disparity reflects a fundamental difference in the semantic nature of content categories. Pornographic content is typically defined by explicit, localized visual cues that enable reliable single-frame classification~\cite{jin2018pornographic}. Violence and crime recognition, however, are inherently context-dependent and require temporal reasoning to distinguish between visually ambiguous actions (e.g., play versus harm)~\cite{wu2020xd-violence}. The compression ablation of Section~\ref{sec:finding2} bears this out. Spatial pooling discards intra-frame detail while leaving the number of sampled frames, and hence the available temporal evidence, entirely unchanged. Removing it therefore replenishes the resource on which localized, single-frame detection depends while doing nothing to relieve the temporal-integration bottleneck that governs Crime and Violence. Accordingly, Porn exhibits an outsized HDR improvement under reduced compression (Table~\ref{tab:ablation_compression}). The sparse frame sampling and limited temporal modeling of current \VideoLLMs make them more reliant on static visual cues, reducing their capacity for fine-grained temporal reasoning. This structural weakness is not addressable by input-level defenses, which operate at the pixel level and do not enhance temporal understanding.

\smallskip
\noindent
\textbf{A temporal defense does not close the gap.}
VideoPure is the one defense in our study built for video rather than for images, and it is therefore the natural candidate to recover the temporally defined categories. It does not. Averaged over all attacks and models, it attains 3.4\% HDR, no better than the 4.2\% mean of the ten image defenses, and its category profile is more skewed, not less. It reaches 7.3\% on Porn against 1.6\% on Crime and 1.4\% on Violence, a $4.8\times$ gap where the image defenses show $2.5\times$.
The reason is that the defense never sees a video. VideoPure enforces temporal consistency by estimating optical flow between adjacent frames, but it is applied, as every defense here is, to the $K=32$ frames the sampler has already selected. Those frames span the full duration of the source video, so consecutive frames in the purified clip may be seconds apart, and the flow field estimated between them does not correspond to any real motion.
Sparse temporal sampling strips the input of the frame-to-frame continuity that video-native defenses are built on, disabling the one class of defense equipped for this modality before it runs. Any defense premised on motion coherence or temporal smoothness inherits the same limitation.

\smallskip
\noindent
\textbf{Implications.}
This disparity has consequences beyond the specific content categories tested.
It suggests that the safety-monitoring capabilities of \VideoLLMs are structurally biased toward content that can be identified from static visual features, while content requiring temporal reasoning, including sequential actions, contextual violence, and escalating threats, is systematically under-detected.
As adversaries become aware of this bias, they can preferentially embed temporally complex harmful content (e.g., instructional crime or contextual harassment) that exploits the models' temporal reasoning deficiency.
Addressing this weakness requires not only improved temporal modeling within \VideoLLMs but also defense mechanisms that are specifically designed to enhance temporal coverage and context retention during the observation pipeline.

%% file: table1.tex
\begin{table*}[!t]
\centering
\resizebox{\textwidth}{!}{%
\begin{tabular}{ll
  rrr
  rrr
  rrr
  rrr
  rrr}
\toprule
& & \multicolumn{3}{c}{L-7B}
& \multicolumn{3}{c}{L-7B-DPO}
& \multicolumn{3}{c}{L-32B}
& \multicolumn{3}{c}{SG4V}
& \multicolumn{3}{c}{VL3} \\
\cmidrule(lr){3-5}\cmidrule(lr){6-8}\cmidrule(lr){9-11}\cmidrule(lr){12-14}\cmidrule(lr){15-17}
Defense & Attack
& Cri & Por & Vio
& Cri & Por & Vio
& Cri & Por & Vio
& Cri & Por & Vio
& Cri & Por & Vio \\
\midrule

\multirow{5}{*}{ComDefend}
& PoisonVID (AKS)  & 7  & 21 & 4  & 0 & 0 & 0  & 10 & 7  & 14 & 0 & 2 & 1 & 9  & 21 & 6 \\
& PoisonVID (FRAG) & 11 & 16 & 19 & 0 & 0 & 0  & \textbf{27} & \textbf{34} & \textbf{29} & 1 & 2 & 3 & 17 & \textbf{28} & 3 \\
& FRA  & 2  & 3  & 1  & 0 & 0 & 0  & 9  & 2  & 12 & 2 & 0 & 0 & 7  & \textbf{31} & 4 \\
& PiP  & 1  & 4  & 2  & 0 & 0 & 0  & 4  & 18 & 5  & 0 & 1 & 0 & 3  & \textbf{25} & 1 \\
& TOA  & 1  & 2  & 1  & 0 & 1 & 0  & 7  & 4  & 2  & 2 & 0 & 2 & 3  & 6  & 5 \\
\midrule

\multirow{5}{*}{DiffPure}
& PoisonVID (AKS)  & 3 & 5 & 5  & 0 & 0 & 0  & 2  & 8  & 10 & 0 & 7 & 0 & 4 & 14 & 4 \\
& PoisonVID (FRAG) & 6 & 5 & 15 & 0 & 0 & 0  & 10 & \textbf{37} & 16 & 0 & 2 & 1 & 9 & 17 & 5 \\
& FRA  & 2 & 2 & 2  & 0 & 0 & 0  & 7  & 3  & 3  & 1 & 2 & 0 & 8 & 12 & 6 \\
& PiP  & 1 & 3 & 1  & 0 & 0 & 0  & 6  & \textbf{27} & 2  & 0 & 1 & 0 & 6 & \textbf{34} & 3 \\
& TOA  & 0 & 1 & 0  & 0 & 0 & 0  & 5  & 3  & 3  & 0 & 0 & 2 & 1 & 4  & 5 \\
\midrule

\multirow{5}{*}{ICC}
& PoisonVID (AKS)  & 1 & 1 & 4 & 1 & 0 & 0 & 4 & 8 & 6 & 0 & 17 & 1 & 7 & \textbf{25} & 7 \\
& PoisonVID (FRAG) & 0 & 2 & 3 & 0 & 0 & 1 & 9 & \textbf{32} & 7 & 1 & 10 & 2 & 11 & \textbf{37} & 5 \\
& FRA              & 0 & 0 & 0 & 0 & 0 & 0 & 5 & 9 & 9 & 0 & 2 & 0 & 6 & \textbf{34} & 9 \\
& PiP              & 1 & 1 & 3 & 0 & 0 & 0 & 1 & 8 & 5 & 0 & 1 & 0 & 4 & 16 & 7 \\
& TOA              & 0 & 3 & 0 & 0 & 0 & 0 & 4 & 9 & 2 & 1 & 1 & 1 & 4 & 7 & 3 \\
\midrule

\multirow{5}{*}{LGS}
& PoisonVID (AKS)  & 2 & 7 & 4  & 0 & 0 & 0  & 3  & 6  & 1  & 1 & 2 & 1 & 3  & 6  & 2 \\
& PoisonVID (FRAG) & 9 & 7 & 17 & 0 & 0 & 0  & 6  & \textbf{32} & 3  & 4 & 6 & 5 & 9  & 14 & 4 \\
& FRA  & 1 & 0 & 1  & 0 & 0 & 0  & 4  & 2  & 3  & 2 & 1 & 1 & 8  & 8  & 6 \\
& PiP  & 1 & 4 & 0  & 0 & 0 & 0  & 3  & 14 & 6  & 0 & 1 & 0 & 10 & 13 & 5 \\
& TOA  & 1 & 3 & 1  & 0 & 1 & 0  & 2  & 5  & 1  & 1 & 0 & 2 & 3  & 8  & 3 \\
\midrule

\multirow{5}{*}{PAD}
& PoisonVID (AKS)  & 3 & 5 & 5  & 0 & 0 & 0  & 6  & 9  & 2  & 1 & 9 & 5 & 2  & 12 & 0 \\
& PoisonVID (FRAG) & 6 & 5 & 15 & 0 & 0 & 0  & 15 & \textbf{41} & 9  & 3 & 8 & 7 & 7  & 21 & 4 \\
& FRA  & 2 & 2 & 2  & 0 & 0 & 0  & 6  & 1  & 9  & 4 & 5 & 3 & 7  & 11 & 2 \\
& PiP  & 1 & 3 & 1  & 0 & 0 & 0  & 4  & \textbf{42} & 5  & 0 & 1 & 1 & 7  & \textbf{36} & 5 \\
& TOA  & 0 & 1 & 0  & 1 & 1 & 0  & 11 & 23 & 6  & 3 & 1 & 4 & 5  & 12 & 4 \\
\midrule

\multirow{5}{*}{PIXD}
& PoisonVID (AKS)  & 1 & 0 & 4 & 3 & 0 & 0 & 3 & 9 & 6 & 0 & 14 & 1 & 5 & 20 & 3 \\
& PoisonVID (FRAG) & 0 & 2 & 3 & 0 & 0 & 0 & 5 & \textbf{36} & 16 & 0 & 8 & 2 & 4 & 23 & 4 \\
& FRA              & 0 & 0 & 0 & 0 & 0 & 0 & 1 & 6 & 5 & 0 & 2 & 0 & 3 & \textbf{27} & 2 \\
& PiP              & 1 & 1 & 3 & 0 & 0 & 0 & 3 & 6 & 4 & 0 & 0 & 0 & 3 & 15 & 5 \\
& TOA              & 3 & 2 & 0 & 0 & 1 & 0 & 4 & 7 & 0 & 0 & 0 & 1 & 1 & 0 & 2 \\
\midrule

\multirow{5}{*}{Quilting}
& PoisonVID (AKS)  & 1 & 3 & 1 & 0 & 0 & 0 & 7 & 10 & 7 & 0 & 0 & 1 & 6 & 16 & 2 \\
& PoisonVID (FRAG) & 3 & 5 & 2 & 0 & 0 & 0 & 12 & \textbf{38} & 15 & 0 & 0 & 0 & 7 & 23 & 2 \\
& FRA              & 6 & 1 & 0 & 0 & 0 & 0 & 7 & 10 & 8 & 0 & 1 & 1 & 0 & 10 & 1 \\
& PiP              & 1 & 0 & 0 & 0 & 1 & 0 & 5 & 8 & 1 & 0 & 0 & 2 & 2 & \textbf{26} & 2 \\
& TOA              & 0 & 1 & 0 & 0 & 1 & 0 & 3 & 0 & 0 & 4 & 1 & 3 & 0 & 4 & 1 \\
\midrule

\multirow{5}{*}{RAND}
& PoisonVID (AKS)  & 1 & 2 & 4 & 0 & 1 & 0 & 4 & 4 & 1 & 0 & 10 & 0 & 7 & 2 & 1 \\
& PoisonVID (FRAG) & 1 & 2 & 1 & 0 & 0 & 0 & 10 & 21 & 11 & 3 & 10 & 3 & 10 & 9 & 2 \\
& FRA              & 1 & 0 & 1 & 0 & 0 & 0 & 4 & 2 & 5 & 0 & 4 & 0 & 5 & 11 & 8 \\
& PiP              & 1 & 1 & 1 & 0 & 0 & 0 & 3 & 6 & 2 & 0 & 2 & 2 & 3 & 8 & 8 \\
& TOA              & 0 & 2 & 2 & 0 & 4 & 0 & 4 & 8 & 0 & 1 & 2 & 2 & 4 & 5 & 3 \\
\midrule

\multirow{5}{*}{SuperRes}
& PoisonVID (AKS)  & 2 & 1 & 1 & 0 & 0 & 0 & 3 & 0 & 4 & 0 & 3 & 2 & 3 & 11 & 1 \\
& PoisonVID (FRAG) & 1 & 4 & 1 & 0 & 0 & 0 & 2 & 12 & 4 & 2 & 2 & 4 & 2 & 17 & 3 \\
& FRA              & 3 & 1 & 0 & 0 & 0 & 0 & 3 & 4 & 1 & 0 & 4 & 2 & 1 & 4 & 0 \\
& PiP              & 1 & 0 & 0 & 0 & 0 & 0 & 1 & 5 & 1 & 0 & 0 & 2 & 1 & 12 & 3 \\
& TOA              & 0 & 1 & 0 & 0 & 2 & 0 & 1 & 2 & 0 & 0 & 1 & 1 & 1 & 2 & 0 \\
\midrule

\multirow{5}{*}{TVM}
& PoisonVID (AKS)  & 3 & 3 & 7 & 0 & 0 & 0 & 11 & 10 & 6 & 0 & 12 & 0 & 12 & 12 & 1 \\
& PoisonVID (FRAG) & 7 & 6 & 5 & 0 & 0 & 0 & 16 & \textbf{26} & 21 & 5 & 7 & 4 & 13 & 17 & 4 \\
& FRA              & 0 & 0 & 2 & 0 & 0 & 0 & 9 & 3 & 8 & 0 & 6 & 0 & 7 & 20 & 1 \\
& PiP              & 0 & 1 & 2 & 0 & 0 & 0 & 3 & 14 & 7 & 0 & 0 & 0 & 2 & 21 & 3 \\
& TOA              & 0 & 2 & 0 & 0 & 0 & 0 & 7 & 7 & 1 & 0 & 1 & 1 & 6 & 2 & 1 \\
\midrule

\multirow{5}{*}{VideoPure}
& PoisonVID (AKS)  & 0 & 3 & 0 & 0 & 0 & 0 & 2 & 3 & 3 & 0 & 4 & 0 & 3 & 12 & 2 \\
& PoisonVID (FRAG) & 0 & 1 & 4 & 0 & 0 & 0 & 3 & \textbf{31} & 6 & 2 & 8 & 1 & 3 & 22 & 6 \\
& FRA              & 0 & 1 & 0 & 0 & 0 & 0 & 2 & 4 & 2 & 0 & 7 & 3 & 2 & 4 & 3 \\
& PiP              & 0 & 0 & 0 & 0 & 0 & 0 & 9 & \textbf{26} & 0 & 0 & 3 & 1 & 2 & \textbf{27} & 1 \\
& TOA              & 0 & 0 & 0 & 0 & 0 & 0 & 9 & 9 & 1 & 2 & 6 & 1 & 2 & 11 & 1 \\
\bottomrule
\end{tabular}
}
\caption{Harmful detection rate (HDR, \%) after applying input-level defenses under observation-level attacks.
Each defense is evaluated on five attack types across three harmful-content categories and five \VideoLLMs.
\textbf{Bold} entries indicate HDR~$\geq$~25\%. Higher is better.}
\label{tab:defenses_combined}
\end{table*}

%% file: anova_fourway.tex
\begin{table}[!t]
\centering
\begin{tabular}{lrrrrr}
\toprule
Term & SS & df & $F$ & $p$ & $\eta^2$ (\%) \\
\midrule
Model & 9013 & 4 & 94.1 & $<10^{-4}$ & 25.4 \\
Category & 3356 & 2 & 70.0 & $<10^{-4}$ & 9.4 \\
Attack & 2812 & 4 & 29.3 & $<10^{-4}$ & 7.9 \\
Defense & 1097 & 10 & 4.6 & $<10^{-4}$ & 3.1 \\
\midrule
Residual (interactions) & 19263 & 804 & --- & --- & 54.2 \\
\bottomrule
\end{tabular}
\caption{Variance decomposition of Table~\ref{tab:defenses_combined} (HDR). Four-way ANOVA over the 825 cells; all interaction terms are pooled into the residual. $\eta^2$ is the share of total HDR variance explained and sums to 100\% across the table.}
\label{tab:anova_fourway}
\end{table}

%% file: siglip.tex
\begin{table}[t]
\centering
\begin{tabular}{lrr}
\toprule
& \multicolumn{2}{c}{$\Delta$ harmful alignment} \\
\cmidrule(l){2-3}
Defense & Omission & Suppression \\
\midrule
ComDefend  & $+1.3\%$  & $-8.1\%$  \\
DiffPure   & $-5.7\%$  & $-18.0\%$ \\
ICC        & $-1.8\%$  & $-2.2\%$  \\
LGS        & $-0.7\%$  & $+0.1\%$  \\
PAD        & $-5.8\%$  & $-4.2\%$  \\
PIXD       & $-2.9\%$  & $-1.8\%$  \\
Quilting   & $-2.6\%$  & $-25.1\%$ \\
RAND       & $-1.0\%$  & $-7.5\%$  \\
SuperRes   & $-0.1\%$  & $-0.2\%$  \\
TVM        & $-2.6\%$  & $-17.7\%$ \\
VideoPure  & $-21.1\%$ & $-29.3\%$ \\
\bottomrule
\end{tabular}
\caption{Change in harmful-concept alignment after each input-level defense,
measured separately for each attack mechanism. Negative values mean harmful content became less
distinguishable.}
\label{tab:siglip}
\end{table}

%% file: ablation_table2.tex
\begin{table}[!t]
\centering
\small
\begin{tabular}{l
  rrr}
\toprule
& \multicolumn{3}{c}{$\Delta$} \\
\cmidrule(lr){2-4}
Model
& Cri & Por & Vio \\
\midrule
L-7B & $+0.0$ & $+4.0$ & $+2.5$ \\
L-7B-DPO & $+6.5$ & $+6.8$ & $+0.2$ \\
L-32B & $+17.0$ & $+14.8$ & $+6.2$ \\
SG4V & $+0.2$ & $+2.2$ & $+4.2$ \\
VL3 & $+4.8$ & $+28.0$ & $-4.0$ \\
\bottomrule
\end{tabular}
\caption{Effect of ablating visual-token compression. Each entry is the change in HDR (percentage
points) when the compression stage is removed, averaged over the four combinations of
DiffPure and PAD with PiP and TOA.}
\label{tab:ablation_compression}
\end{table}

%% file: discussion.tex
\section{Discussion}
\label{sec:discussion}

Section~\ref{sec:experiment} established four empirical findings and, for each, the mechanism that produces it. Input-level defenses provide limited protection because they act only on already-sampled frames. This failure persists even when harmful content is present in every frame, because token compression discards it before the language model sees it. Model architecture dominates defense effectiveness, and detection rates vary drastically across content categories.
We now turn from why the tested defenses fail to what a defense that does not fail would have to do.

\subsection{Toward System-Level Robustness}
\label{sec:system_robustness}

Our study does not argue that input-level robustness techniques are unimportant; rather, it clarifies their scope.
They can improve robustness to perturbations \emph{conditional on the model having access to the safety-relevant visual content}, but they do not guarantee robustness when the observation pipeline, spanning frame sampling, token compression, and modality fusion, prevents the model from perceiving that content in the first place.
This motivates future defense directions that address the full observation pipeline:
\begin{itemize}
    \item \textbf{Sampling-aware coverage guarantees.} Strategies that ensure safety-relevant temporal segments are represented in the sampled subset, regardless of adversarial manipulation of relevance scores.
    \item \textbf{Token-level preservation.} Mechanisms that preserve localized and low-contrast visual features during compression, preventing safety-relevant signals from being discarded by token budgeting. Our compression ablation (Section~\ref{sec:finding2}) demonstrates that this direction is viable, since restoring the token budget more than doubles pooled HDR.
    \item \textbf{Modality-balanced fusion.} Architectures that prevent systematic down-weighting of attenuated visual signals during multimodal integration, ensuring that weakened but present harmful content still influences the model's output.
    \item \textbf{Observation verification.} Post-hoc mechanisms that detect when sampling or compression is likely to have discarded critical content, triggering re-examination of the input.
\end{itemize}
More broadly, our results point to a gap between input-level robustness and system-level robustness in multimodal architectures with structured observation pipelines.

\subsection{Limitations}
\label{sec:limitations}

Our evaluation focuses on representative \VideoLLMs, sampling mechanisms, and observation-level attacks that are currently available, and on a detection-style safety task with a standardized response-to-label mapping.
While this setting enables controlled comparisons, other tasks (e.g., long-form summarization and multi-turn QA) may exhibit additional failure modes.
Moreover, we adapt input-level defenses to the video setting by applying them to sampled frames; alternative integration strategies (e.g., operating on latent features or jointly with sampling) may yield different trade-offs.
Finally, our findings characterize transferability under the tested configurations and should not be interpreted as a definitive impossibility result for all future defenses.

%% file: conclusion.tex
\section{Conclusion}
\label{sec:conclusion}

This paper studies the adversarial robustness of \VideoLLMs under observation-level attacks, which operate on the observation pipeline through omission and suppression rather than through bounded input perturbations.
Because practical \VideoLLMs rely on frame sampling and token budgeting, their predictions are conditioned on partial observation, which enables observation-level attacks that either keep harmful evidence out of inference or exploit its being discarded within it.
We present \fName, a controlled framework for systematically evaluating whether input-level defenses transfer to this threat model under fixed sampling configurations. Across five \VideoLLMs, five observation-level attacks, and eleven representative input-level defenses spanning five families, we find that input-level defenses provide limited and inconsistent mitigation. Critically, defenses fail even when human-visible harmful signals persist in every sampled frame. Defense effectiveness is dominated by model architecture rather than by the defense method, and detection rates vary drastically across content categories, exposing weaknesses in the temporal reasoning of \VideoLLMs.
These results underscore the need to move from input-level robustness toward system-level robustness, spanning sampling-aware coverage guarantees, token-level preservation of safety-relevant features, and modality-balanced fusion, for securing \VideoLLMs in safety-critical deployments.